\documentclass{article}
\usepackage{PRIMEarxiv}

\usepackage[utf8]{inputenc} 

\usepackage{hyperref}       
\usepackage{url}            
\usepackage{booktabs}       
\usepackage{amsfonts}       
\usepackage{nicefrac}       
\usepackage{microtype}      
\usepackage{lipsum}
\usepackage{fancyhdr}       
\usepackage{graphicx}       

\usepackage{tikz}
\usepackage{comment}
\usepackage{amsmath,amssymb} 
\usepackage{algorithm,algorithmic}
\usepackage{color}
\usepackage{booktabs}
\usepackage{makecell}
\usepackage{multirow}
\usepackage{graphics}
\usepackage{caption}
\usepackage{subcaption}
\usepackage{todonotes}

\renewcommand{\algorithmiccomment}[1]{\bgroup\hfill//~#1\egroup}

\title{FaceMap: Towards Unsupervised Face Clustering via Map Equation}

\author{Xiaotian Yu\thanks{ denotes equal contributions}, Yifan Yang$^*$, Aibo Wang, Ling Xing, Hanling Yi\\
Intellifusion Technology Ltd., Shenzhen \\
\texttt{\{yu.xiaotian,wang.aibo,xing.ling\}@intellif.com}\\
\texttt{yifan.yang.cn@gmail.com; hanling.cuhk@gmail.com}
\And Guangming Lu \\
Harbin Institute of Technology, Shenzhen \\
\texttt{luguangm@hit.edu.cn}
\And Xiaoyu Wang \\
    Intellifusion Technology Ltd., Shenzhen \\
    The Chinese University of Hong Kong, Shenzhen\\
    \texttt{wang.xiaoyu@intellif.com}}

\begin{document}

\maketitle

\begin{abstract}
Face clustering is an essential task in computer vision due to the explosion of related applications such as augmented reality or photo album management. 
The main challenge of this task lies in the imperfectness of similarities among image feature representations. Given an existing feature extraction model, it is still an unresolved problem that how can the inherent characteristics of similarities of unlabelled images be leveraged  to improve the clustering performance.
Motivated by answering the question, we develop an effective unsupervised method, named as FaceMap, 
by formulating face clustering as a process of non-overlapping community detection, and minimizing the entropy of information flows on a network of images. The entropy is denoted by the map equation and its minimum represents the least description of paths among images in expectation.  
Inspired by observations on the ranked transition probabilities in the affinity graph constructed from facial images, we develop an outlier detection strategy to adaptively adjust transition probabilities among images. Experiments with ablation studies demonstrate that FaceMap
significantly outperforms existing methods and achieves new state-of-the-arts on three popular large-scale datasets for face clustering, \emph{e.g.}, an absolute improvement of more than $10\%$ and $4\%$ comparing with prior unsupervised and supervised methods respectively in terms of average of Pairwise F-score. Our code is publicly available on github~\footnote{https://github.com/bd-if/FaceMap}. 
\end{abstract}


\section{Introduction}
Face clustering has received considerable attention over the last decade due to the advancement of deep learning in face recognition.
It has vast applications for personal photo management and entertainment such as augmented reality effects created by face morphing from multiple photos~\cite{scherhag2019face}, face tagging in albums~\cite{zhu2011rank}, video analysis~\cite{cao2014robust}, \emph{etc}. It can also be utilized to effectively annotate large-scale facial image datasets~\cite{tian2007face,wang2019linkage} or perform label noise cleansing~\cite{zhang2020global}.

Face clustering aims at grouping facial images with the same identity into one cluster, while simultaneously discriminating different identities via different cluster labels~\cite{shen2021structure,wang2019linkage,xing2021learning,yang2020learning,yang2019learning,zhang2020global}.
The task is easy if we have a perfect face feature extractor, \emph{i.e.}, similarities of face features belonging to the same identity are always much higher than those from different identities. However, it is always not true for real world scenarios.
Given a pre-trained feature model and a set of images, a common practice of face clustering is to firstly construct an affinity graph based on image similarities with $k$ nearest neighbors ($k$NN), where an image is a node and the cosine similarity between image features is the weight of an edge. The resulting graph is typically noisy with incorrect connections or missing edges, which may lead to performance degradation for subsequent clustering tasks through graph partitioning, \emph{etc}.

Prior studies with supervised methods train a model with annotated samples to reduce the noise in the affinity graph, and achieve the state-of-the-arts in face clustering~\cite{shen2021structure}. However, the expensive cost of label annotations and the huge hyper-parameter tuning hinder the applications of supervised methods in real-world scenarios. Thus it is worth inventing an unsupervised method that is capable of working with imperfect facial feature representations and achieves outstanding clustering performance. 


\begin{figure}[!t]
    \centering
    \begin{subfigure}{.5\textwidth}
    \includegraphics[width=\linewidth]{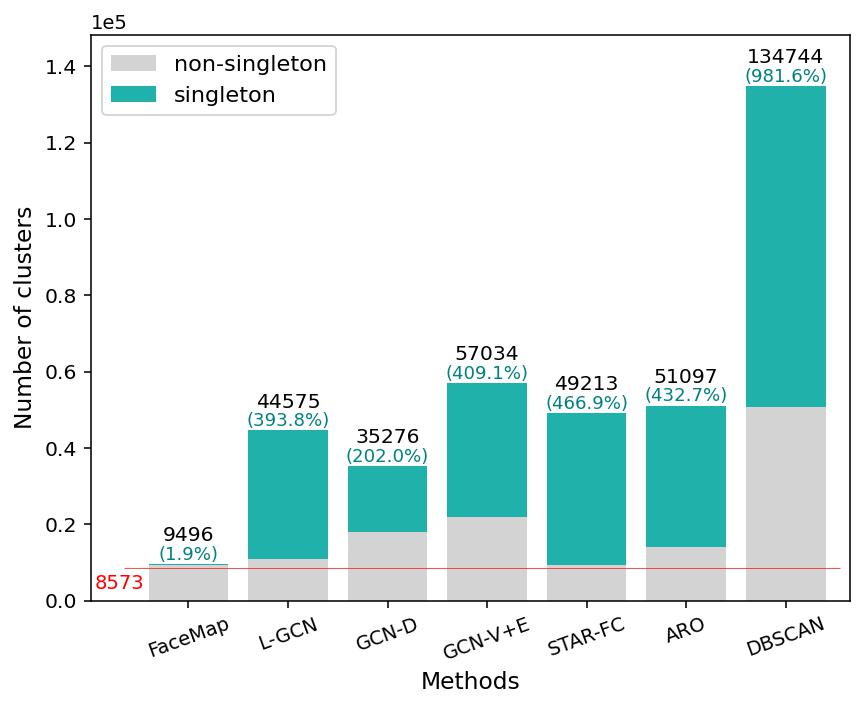}
    \caption{Part1(0.58M) in MS1M}
    \end{subfigure}%
    \begin{subfigure}{.5\textwidth}
    \includegraphics[width=\linewidth]{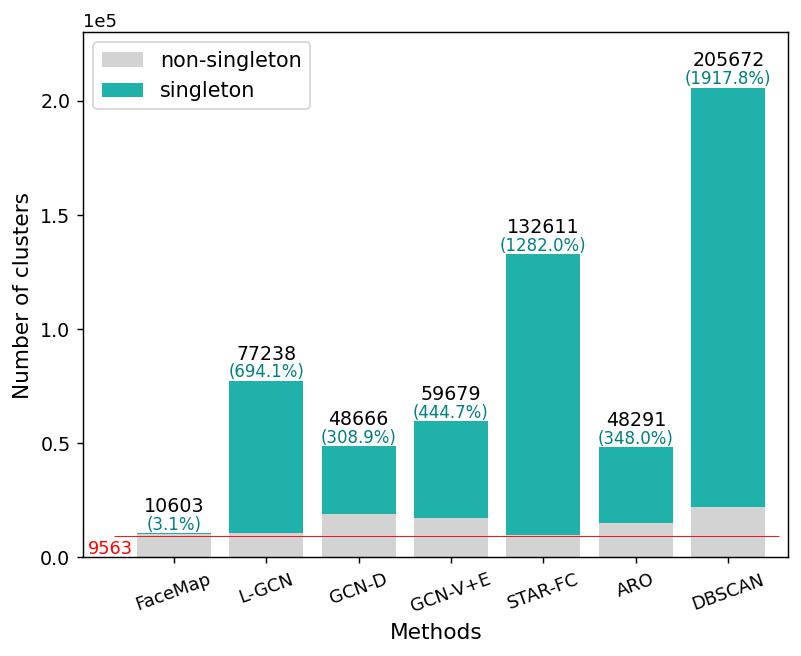}
    \caption{CASIA}
    \end{subfigure}
    \caption{The comparison on the number of predicted clusters among different clustering methods for datasets of MS1M~\cite{guo2016ms} and CASIA~\cite{yi2014casiawebface}. The red line with number denotes the true number of identities in the dataset. For each method, the total number of predicted clusters and the ratio of singleton cluster are shown. Singleton cluster means that the size of a predicted cluster is one. We clearly observe that, for prior methods, the number of clusters exceeds a lot the true number of identities, and the ratio of singleton cluster is large. Our FaceMap shows superiority in terms of the number of predicted clusters.}
    \label{fig:number-of-cluster}
\end{figure}

In addition, performance of face clustering methods is usually evaluated based on Pairwise F-score ($F_\text{P}$)~\cite{shi2018face} and BCubed F-score ($F_\text{B}$)~\cite{amigo2009comparison}. The two traditional metrics are biased toward large-size clusters~\cite{amigo2009comparison,moreno2015adapted}, which grossly neglect the negative impact of 
incorrect partitions on small-size clusters. 
Those clusters create lots of burdens for subsequent applications because they misinformed the true number of clusters. Moreover, $F_\text{P}$ and $F_\text{B}$ do not have sufficient characterization on the difference between the number of predicted clusters and the true number of identities.
However, the corresponding information could be critical for applications that need to understand the true number of unique persons. 


For further illustrations, we show the comparison on the number of predicted clusters among different methods in Fig.~\ref{fig:number-of-cluster}. In prior methods, we clearly observe that the number of resulting clusters could be 10$\times$ larger than the true number of identities, and singleton clusters could be the dominant portion. It was pointed out in~\cite{wang2019linkage} that singleton clusters usually contain hard samples, which means the illustrated algorithms actually turn hard examples into singleton clusters to improve performance.


To this end, we study the face clustering problem from two perspectives. On one hand, we formulate unsupervised face clustering as a process of non-overlapping community detection based on the map equation~\cite{rosvall2009map}, where each identity is an underlying community and the map equation characterizes the entropy of paths by a random walker travelling over the affinity graph. The minimum value of the map equation represents the least description of paths among images in expectation.
The transition probability from a source node to a target node in the affinity graph is the normalized similarity between the source node and the target node, where the normalization is over all connected target nodes from the source node.
To address the challenge of imperfect feature extractors, we develop a strategy of outlier detection (OD) to adaptively adjust transition probabilities of a given affinity graph. Our proposed FaceMap method, equipped with the OD module, minimizes the map equation and performs high quality face clustering without using any labeled data. High quality face clustering not only achieves outstanding performance on all images, but also has satisfactory results on identity-level images.  On the other hand, we introduce three new metrics for evaluating face clustering performance which contains key implications about clustering quality in real applications. The new metrics measure identity-level quality, which take incorrect partitions of small-size clusters into account and measure the discrepancy between the number of predicted clusters and the true identity numbers. We also present corresponding studies of state-of-the-art methods with the new metrics.

In this paper, we show that a dedicatedly-designed unsupervised method has the capability of outperforming all the existing state-of-the-art methods.
To our best knowledge, face clustering has not been investigated from the viewpoint of non-overlapping community detection. Moreover, there is little study that the map equation is applied into computer vision with significant performance. In summary, we make the following contributions.
\begin{itemize}
    \item To the best of our knowledge, we are the first to formulate face clustering as community detection with the map equation. We propose an effective method that adaptively adjusts the distribution of transition probabilities in the affinity graph for face clustering.
    \item We illustrate the limitations of the traditional metrics for face clustering. For a comprehensive comparison among methods in clustering facial images, we design three metrics for evaluating the identity-level quality.
    \item The unsupervised FaceMap significantly outperforms the prior unsupervised and supervised methods, and achieves new state-of-the-arts in light of traditional metrics on three large-scale datasets.
    We also show the superiority of our method in terms of identity-level quality via new metrics. 
\end{itemize}

\section{Problem Definition, Metrics and Related Works}

In this section, we first present the problem definition of face clustering. Then, we show the limitations of traditional metrics ($F_\text{P}$ and $F_\text{B}$) in this task. We thereby design three new metrics as complementary evaluations from identity-level quality perspective.
Finally, we present related works on this topic. 

\subsection{Problem Definition}

In this paper, we study the problem of face clustering with images. In particular, the input of face clustering is a set of images $X= \{x_i\}^S_{i=1}$, where  $x_i\in R^d$ denotes the facial feature of  an image and $d$ is the dimension of the feature. Face clustering requires an algorithm to produce a set of predicted labels $Y=\{y_i\}^S_{i=1}$, where $y_i$ is the predicted label to each image $x_i$ with $y_i\in \{1,2,\cdots,N\}$, and $N$ denoting the number of predicted clusters. Note that $N$ is unknown and should be determined during clustering. Images with the same predicted label form a cluster. The set of true labels is defined as  $Y^*_i=\{y^*_i\}^S_{i=1}$ where $y^*_i=\{1,\cdots,N^*\}$ with $N^*$ denoting the true number of identities. Predicted clusters are denoted by $C=\{C_j\}^N_{j=1}$ with $C_j=\{x_i|y_i=j,\forall x_i \in X\}$, and the true identities are denoted by a set $T=\{T_l\}^{N^*}_{l=1}$ with $T_l=\{x_i|y^*_i=l,\forall x_i\in X\}$. Singleton clusters should be a subset of $C$,  and we denote the number of incorrectly predicted singleton clusters as $N_\text{S}$.



\subsection{Metrics for Face Clustering}\label{sec:metrics}

\textbf{1) Traditional metrics.} $F_\text{P}$ and $F_\text{B}$ are two widely used metrics to evaluate clustering performance. Basically, the performance of $F_\text{P}$ and $F_\text{B}$ is dominated by the large-size clusters~\cite{amigo2009comparison,moreno2015adapted}. Note that we omit the metric of NMI in prior works~\cite{shen2021structure,wang2019linkage,xing2021learning,yang2020learning,yang2019learning,zhang2020global} due to its tendency to choose the results with large number of clusters~\cite{amelio2015normalized}.

\begin{figure}[!t]
    \centering
    \includegraphics[width=\linewidth]{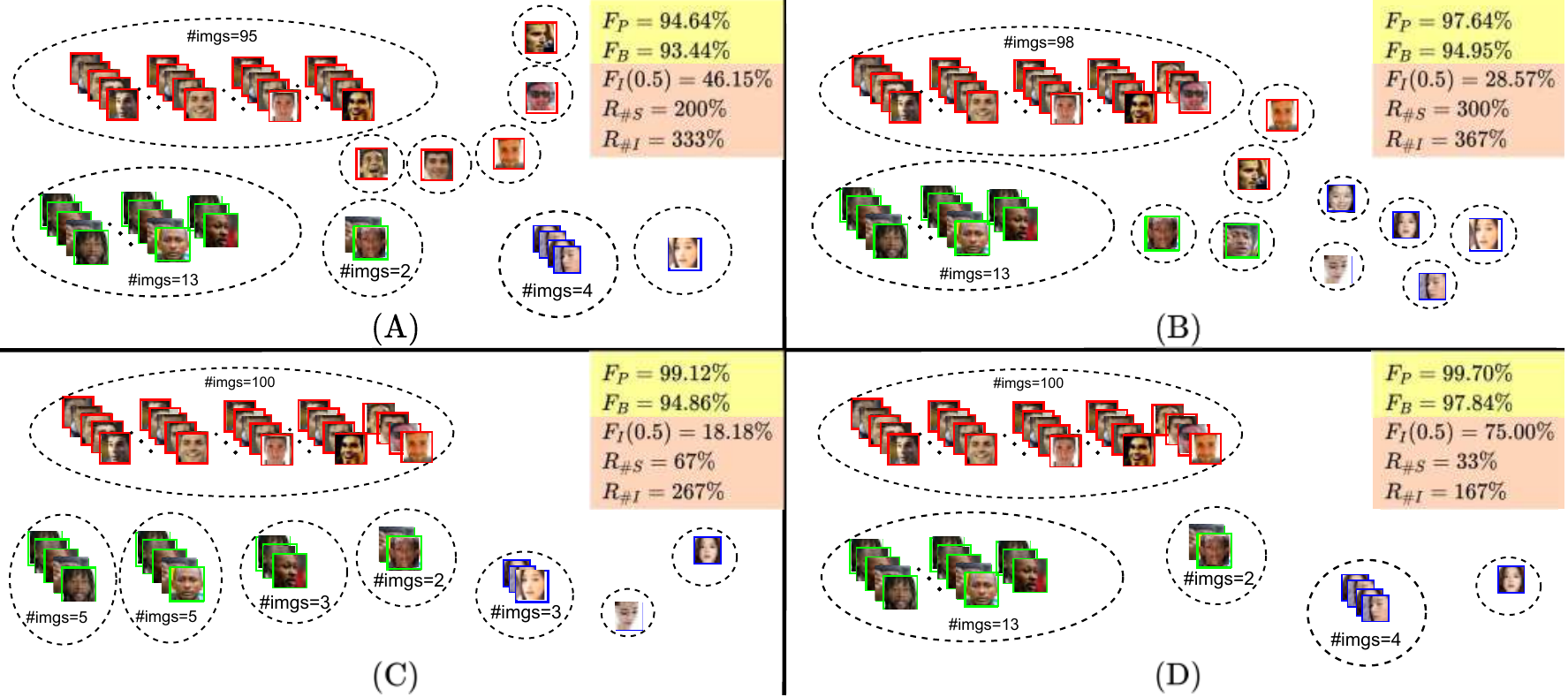}
    \caption{We illustrate the limitations of $F_\text{P}$ and $F_\text{B}$ with four examples of clustering a set of images, denoted by (A)-(D). Face images with bounding boxes via the same color belong to the same identity, \emph{i.e.}, 3 identities in examples. Each predicted cluster is represented by a dashed-line ellipse. The results by traditional metrics and new metrics are respectively shown in yellow and pink blocks. We observe that the images of the identity in blue bounding boxes are failed to group together in (B), leading to more singleton clusters compared to (A), however $F_\text{P}$ and $F_\text{B}$ of (B) are higher than those of (A). Similarly, $F_\text{P}$ and $F_\text{B}$ of (C) are higher than those of (A), but the images of two identities in (C) are heavily split. By contrast, the new metric $F_\text{I}$ is sensitive to the small-size clusters with false identities, and correctly discriminates the results in (A)-(C). Similar cases are found in $R_{\#\text{S}}$ and $R_{\#\text{I}}$. For a complete comparison, good performance should be reported not only by $F_\text{P}$ and $F_\text{B}$, but also by $F_\text{I}$, $R_{\#\text{S}}$ and $R_{\#\text{I}}$, as shown in (D).}
    \label{fig:metrics}
\end{figure}

The limitations of the above two metrics are illustrated in Fig.~\ref{fig:metrics}, where we show four examples (\emph{i.e.}, (A)-(D) in the figure) of clustering results with three identities. We observe that the images of the identity in blue bounding boxes are failed to group together in (B), leading to more singleton clusters compared to (A), however $F_\text{P}$ and $F_\text{B}$ of (B) are higher than those of (A). Similarly, $F_\text{P}$ and $F_\text{B}$ of (C) are higher than those of (A), but the images of two identities in (C) are heavily split. Thus higher $F_\text{P}$ and $F_\text{B}$ do not imply better clustering results, especially from identity-level quality perspective.


\textbf{2)  Proposed new additional metrics.} We design three new metrics, named as Ratio of Identity Number ($R_{\#\text{I}}$), Ratio of Singleton Cluster Number ($R_{\#\text{S}}$), and Identity F-score ($F_{\text{I}}$), for complementary evaluation of face clustering.

$R_{\#\text{I}}$ is the ratio of the number of predicted clusters to the true identity number: $R_{\#I} = N/N^*\times 100\%$. $R_{\#\text{I}}$ is closer to 100\%, the better.

$R_{\#\text{S}}$ is the ratio of the number of incorrectly predicted singleton clusters to the true identity number: $R_{\#S} = N_S/N^*\times 100\%$. $R_{\#\text{S}}$ is smaller, the better.

$F_{\text{I}}$ is inspired from the metric IDF1 in the evaluation of multiple objects tracking~\cite{ristani2016performance}. To evaluate the performance of imbalanced data, $F_{\text{I}}$ measures the degree of alignment between predicted clusters and identities. 
Given an associated pair of sets $APair(j,l) = (C_j,T_l)$, where $j\in \{1,\cdots,N\}$ and $l\in \{1,\cdots,N^*\}$, we can calculate precision and recall on $APair(j,l)$, \emph{i.e.}, $Pre(j,l) = |C_j \cap T_l|/|C_j|$ and $Rec(j,l)=|C_j \cap T_l|/|T_l|$. Given a quality threshold $\theta>0$, we define optimal associated pairs by $OAPair(j,l,\theta)$, where $Pre(j,l)>\theta$ and $Rec(j,l)>\theta$. We may evaluate the identity-level quality by calculating F-score based on the optimal associated pairs, \emph{i.e.},
\begin{equation}
    F_I(\theta) = 2\cdot\frac{Pre(\theta)\cdot Rec(\theta)}{Pre(\theta) + Rec(\theta)},
\end{equation}
where $Pre(\theta)=|OAPair(j,l,\theta)|/N$ and $Rec(\theta)=|OAPair(j,l,\theta)|/N^*$. The degree of identity-level quality can be controlled by $\theta$ according to the purpose of applications in practice, and $\theta\in[0.5, 1)$.

The effectiveness of the new metrics is shown in Fig.~\ref{fig:metrics}. The new metric $F_\text{I}$ is sensitive to the small-size clusters with false identities, and correctly discriminates the results in (A)-(C). Similar cases are found in $R_{\#\text{S}}$ and $R_{\#\text{I}}$. By comparing the four examples in Fig.~\ref{fig:metrics}, we clearly observe that the combined evaluation metrics better illustrate the performance of face clustering. 



\subsection{Related Works}
In this subsection, we present two streams of studies on face clustering in terms of whether annotated samples are used or not, \emph{i.e.}, unsupervised methods and supervised methods. We also give a brief discussion on the map equation~\cite{rosvall2009map}.

\textbf{1) Unsupervised methods.}
Clustering methods without help from annotated samples have been investigated for facial images~\cite{gan2020data,jain1999data,xu2005survey}. However, the performance of traditional clustering methods, such as K-means~\cite{likas2003global} and DBSCAN~\cite{ester1996density}, is not satisfactory, especially on large-scale datasets, which has been reported in previous studies~\cite{bijl2018comparison,shen2021structure,yang2020learning,yang2019learning}.
The main reason is that these traditional methods resort to simplistic assumptions on data distributions. For example, K-means needs to pre-set the number of predicted clusters and implicitly assumes that the numbers of samples for different clusters are roughly balanced. Besides, certain techniques, \emph{e.g.}, agglomerative hierarchical clustering (HAC)~\cite{sibson1973slink}, have been developed to partition facial data with complex distributions. Lin et al. proposed proximity-aware hierarchical clustering in~\cite{lin2017proximity} and density-aware clustering in~\cite{lin2018deep} for face clustering. Zhu et al.~\cite{zhu2011rank} proposed rank-order distance for clustering images, and demonstrated its ability in filtering outliers. The methods in~\cite{lin2018deep,lin2017proximity,zhu2011rank} are hard to scale to  large datasets due to complexity. An approximation version of rank-order distance is proposed in~\cite{otto2017clustering}, which is termed as ARO and is capable of clustering faces at millions scale. However, the performance of ARO decreases rapidly when the data scale increases and is highly influenced by hyper-parameters.


\textbf{2) Supervised methods.}
The recent advanced supervised-based methods for face clustering are built with deep learning models. The most studied branch is based on Graph Convolutional Network (GCN).
GCN is trained to produce a similarity measure between two facial features with the whole graph structure and thus can be used to correct the noisy affinity graph in face clustering. Clearly, GCN-based methods require annotated samples to learn graph structures. 

There are several techniques proposed for training the GCN model. The first is link prediction, which means that GCN models are trained to predict a link between two nodes in the affinity graph~\cite{wang2019linkage}. The second is node confidence estimation~\cite{yang2020learning}. Each node is estimated with confidence of belonging to a cluster by edge connectivity. The third is sub-graph structure learning~\cite{shen2021structure,yang2019learning}.


Other supervised methods for face clustering include multi-layer perceptron and transformer. Liu et al.~\cite{liu2021learn} select face pairs based on neighborhood density and train a classification model with annotated data. 
Nguyen et al.~\cite{nguyen2021clusformer} formulate 
face clustering as a classification of pairwise relationships in the sequences of $k$NN. 

\textbf{3) Map equation.} Map equation is entropy based on information flows of a random walk on networks, and it can be solved by Infomap algorithm~\cite{rosvall2009map}. In map equation, information flows refer to sequential transitions among nodes in the graph, which are optimized based on the transition probability matrix. The map equation and its solver Infomap are originally proposed for the analysis of complex systems, especially in physics~\cite{lambiotte2019networks,Rosvall1118,rosvall2010mapping} and biology~\cite{edler2017infomap}. Traditional applications of Infomap include structure learning on human brain functional networks~\cite{nicolini2016modular}. 

\section{Method}

In this section, we first formulate face clustering as community detection. Then, we propose an effective unsupervised face clustering framework called FaceMap. Equipped with an OD module, it is able to conduct high quality face clustering.

\subsection{Face Clustering as Community Detection}

Non-overlapping community detection is a process that takes a graph $G$ as input, and produces a set of communities $M$, which are clusters of nodes in the graph, as output by optimizing a particular objective function, \emph{i.e.},
\begin{equation}
    \arg\min_M f(G,M),\label{eqn:general-objective}
\end{equation}
where $M=\{m_i|m_i \cap m_j = \emptyset~\text{with}~i\neq j, m_i\neq \emptyset, 1\leq i,j \leq |M| \}$ and $f$ is an objective function. 
The solution $M$ corresponds to $C$ in face clustering.

Given a set of facial features, we calculate a directed affinity graph $A$ based on $k$NN, where $A\in R^{S\times S}$. For large-scale datasets,  we usually have $S \gg k$ and thus $A$ is highly sparse. The affinity graph characterizes the information flows within images, which are inherently quantified by similarities. The transition probability matrix denoted by $P$ is obtained by row normalization of $A$.


We adopt map equation in~\cite{rosvall2009map}, which represents the entropy of information flows,  as  the  objective  function  in  Eq.~(\ref{eqn:general-objective}). Thus, the map equation for face clustering can be formulated as:

\begin{equation}
    \arg\min_{N,Y} L(P,N,Y) =  - q_\curvearrowright \sum^N_{i=1} \frac{q_{i\curvearrowright}}{q_\curvearrowright}  \log \frac{q_{i\curvearrowright}}{q_\curvearrowright} - \sum^N_{i=1} p_{i\circlearrowright}\left(\sum^N_{i=1} \frac{q_{i\curvearrowright}}{p_{i\circlearrowright}} \log \frac{q_{i\curvearrowright}}{p_{i\circlearrowright}}+\sum_{\alpha \in i} \frac{p_\alpha}{p_{i\circlearrowright}} \log \frac{p_\alpha}{p_{i\circlearrowright}}\right),\label{eqn:face-loss}
\end{equation}
where $Y$ is predicted labels, and $N$ is the predicted number of identities for facial images. $q_\curvearrowright=\sum^N_{i=1} q_{i\curvearrowright}$  and $p_{i\circlearrowright}=q_{i\curvearrowright}+\sum_{\alpha \in i} p_\alpha$ with $\alpha \in i$ denoting over all nodes $\alpha$ in cluster $i$, $p_{\alpha}$ being the ergodic node visit frequency at node $\alpha$ by a random walk, and $q_{i\curvearrowright}=\sum_{\alpha \in i}\sum_{\beta \not\in i} p_\alpha P(\alpha, \beta)$ denoting the per step probability that the random walker exits cluster $i$. Here, $P(\alpha, \beta)$ is the outgoing probability from node $\alpha$  to  $\beta$ in the transition probability matrix.


\begin{figure}[!th]
\centering
\includegraphics[width=\textwidth]{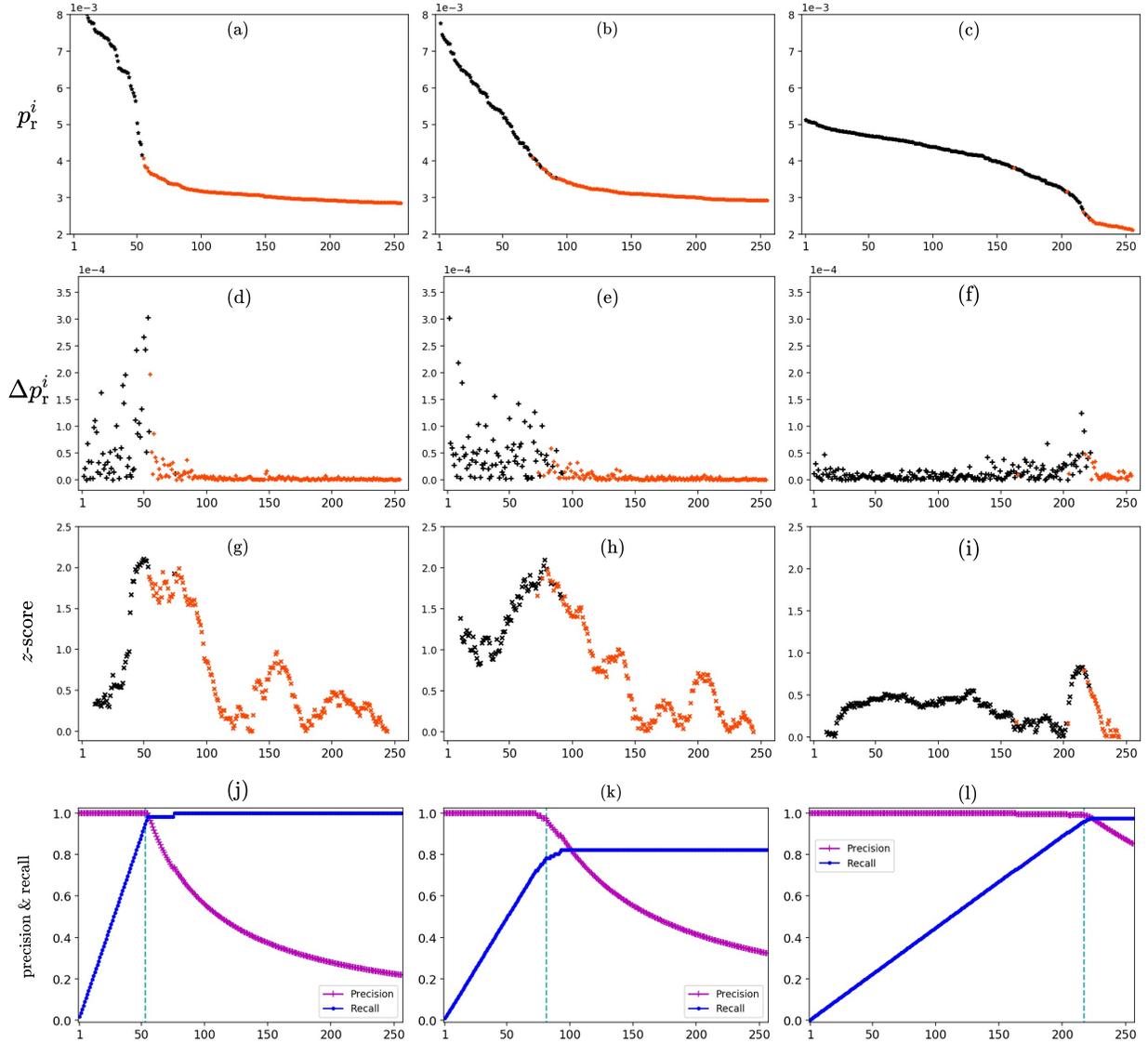}
\caption{The observations of data inherent characteristics are shown with three illustrative examples on CASIA. Each example is a group of four sub-figures in the same column. Given an image $x_i$, $p_{\text{r}}^i$ is the ranked sequence by descending order of non-zero transition probabilities, \emph{i.e.}, (a)-(c). $\Delta p_{\text{r}}^i$ is the first-order difference of $p_{\text{r}}^i$, \emph{i.e.}, (d)-(f). The black and red points represent the images with the same identity of $x_i$ and the images with  different identities of $x_i$ respectively. Given a sliding window $\omega$ by starting from right to left, we estimate $z$-score of average $\Delta p_{\text{r}}^i$ over the samples in the window $\omega$, which is shown in (g)-(i). The estimated switch point is the middle position of $\omega$ with the maximum $z$-score. We show the precision and recall defined in Eqs.~(\ref{eqn:pre}) and (\ref{eqn:rec}) in (j)-(l).}
\label{fig:obscasia}
\end{figure}

In Eq.~(\ref{eqn:face-loss}), we have the following intuitive understandings.
\begin{enumerate}
    \item $q_\curvearrowright$ is the probability of a random walker travelling among different clusters. $q_{i\curvearrowright}$ is the probability of a random walker jumping from the cluster $i$ to other clusters. $p_{i\circlearrowright}$ is the probability of a random walker travelling in the cluster $i$. $p_\alpha$ is the probability of a random walker visiting the node $\alpha$.
    \item The vanilla design of the map equation applies coding theory into network discovery, which compresses the description of information flows on networks. For face clustering, by minimizing the entropy in Eq.~(\ref{eqn:face-loss}), we encode paths among images with $N$ clusters via minimal length of descriptions.
    \item The above map equation has two terms. The first term describes the entropy of random walk travelling among different clusters, and the second term characterizes the entropy of random walk within a cluster. Besides, the number of clusters is also optimized based on the map equation. 
\end{enumerate}

\subsection{Key Observations}
In Fig.~\ref{fig:obscasia}, we give three examples, each of which is demonstrated by a column, to illustrate our observations. Given an image $x_i$, its transition probabilities to other images can be obtained from $P(i,:)$. A rank sequence in a descending order of non-zero values of $P(i,:)$ is denoted as $p_\text{r}^i$, which is shown in (a)-(c) in Fig.~\ref{fig:obscasia}. For $x_i$, the black and red points represent the images with the same identity of $x_i$ (denoted as positive samples) and the images with different identities of $x_i$ (denoted as negative samples) respectively. Clearly, with the view from the left to the right, the black points mainly locate at the head region while the red points locate at the tail region. Besides, there exists a mixed region from the head region to the tail region, \emph{e.g.}, from 50 to 100 in (b) of Fig.~\ref{fig:obscasia}. From the viewpoint of signal processing and clarifying the inherent characteristic of $p_\text{r}^i$, we calculate the first-order difference of $p_\text{r}^i$ denoted by $\Delta p_\text{r}^i$, which is shown in (d)-(f) in Fig.~\ref{fig:obscasia}. For $p_\text{r}^i$ and $\Delta p_\text{r}^i$, we have the following observations.

\begin{figure}[!t]
    \centering
    \begin{subfigure}{.5\textwidth}
    \includegraphics[width=.9\linewidth]{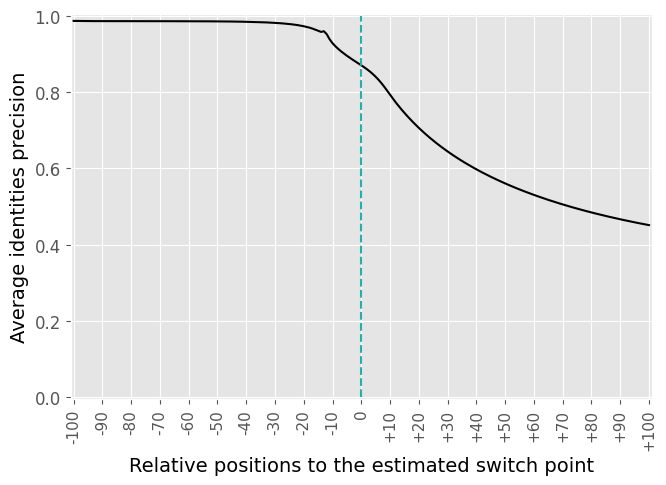}
    \caption{}
    \label{fig:precision}
    \end{subfigure}%
    \begin{subfigure}{.5\textwidth}
    \includegraphics[width=.9\linewidth]{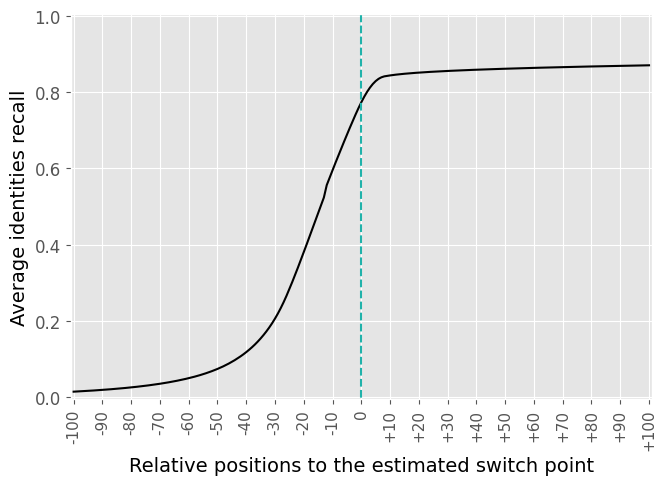}
    \caption{}
     \label{fig:recall}
    \end{subfigure}%
    \caption{The average precision and recall of each relative position to the estimated switch position over all identities.}
    \label{fig:statscasia}
\end{figure}



\textbf{Observation 1.}  For $p_\text{r}^i$, there is no consistent behaviour on the slopes at the head region over all $i\in \{1,\cdots,S\}$. For example, the head region of $p_\text{r}^i$ for (a) is steeper, compared to that for (b) and (c) in Fig.~\ref{fig:obscasia}.

\textbf{Observation 2.}  For $p_\text{r}^i$, there is a stable trend on the slopes at the tail region over all $i\in \{1,\cdots,S\}$, \emph{i.e.}, the slope of $p_\text{r}^i$ is convergent. In other words, the values of $\Delta p_\text{r}^i$ is nearly zero at the tail region.

\textbf{Observation 3.}  The switch position between black and red points, which is usually located in the mixed region of $p_\text{r}^i$, may be easily detected by $\Delta p_\text{r}^i$ with the view from the right to the left.

Based on the above observations, we clearly find that there is no general fixed value of $k$ in nearest neighbor search, or fixed threshold of transition probability, to locate the switch point for any $i\in \{1,\cdots,S\}$.  We present the following assumptions related to data characteristics of $P$.

\textbf{Assumption 1.}  For each $x_i$, we assume that the images with the same identity of $x_i$ and the images with different identities of $x_i$ are roughly separated by a switch point. 

\textbf{Assumption 2.} We assume that the values in $\Delta p^i_\text{r}$ at the right-hand side of the switch point form a stable distribution, and the switch point is an outlier to the stable distribution.




We give a brief discussion on Assumptions 1 and 2. In general, there roughly exists a switch point for each image with the ranked transition probabilities in practice due to the effectiveness of a pre-trained feature model. Note that the switch point should lie in the mixed region. This implies that a few black points might locate at the tail region, which in fact results from the imperfect feature model. The switch point is an outlier to the stable distribution implying that there is a dramatic change of probabilities between the head and tail regions.

To verify the effectiveness of the switch point, we further discuss the data characteristics as follows. For the sequence of $p_\text{r}^i$, given a specific position $t\in [1,K]$ over $p_\text{r}^i$, we may calculate the precision and recall of positive samples with respect to $t$, \emph{i.e.},
\begin{equation}
    Pre(t) = N_t/t,\label{eqn:pre}
\end{equation}
\begin{equation}
    Rec(t) = N_t/N_i,\label{eqn:rec}
\end{equation}
where $N_t$ is the number of positive samples over $[1,t]$ and $N_i$ is the total number of positive samples for image $x_i$.

Given image $x_i$, we may demonstrate the curves of precision and recall over $[1, K]$, with three examples shown in sub-figures (j)-(l) of Fig.~\ref{fig:obscasia}. We clearly find that the switch point locates at the position with high precision and recall for the three examples. 
The average curves of precision and recall over all the identities are shown in Fig.~\ref{fig:statscasia}.
In Fig.~\ref{fig:precision}, we can see that  over $80\%$ nodes on the left side of the estimated switch position  are positive samples. This indicates that majority of edges in the adjusted affinity graph are connected with two images in the same identity, thus their transition probability should be added more weights by row normalization of the transition probability matrix.
In Fig.~\ref{fig:recall}, we observe that nearly $80\%$ positive samples can be included on the left side of the estimated switch position. 
This indicates that $80\%$ of the positive connections are included in the affinity graph. 
As a result, the adjusted affinity graph has dense connections among the nodes within the same identity and sparse connections among the nodes from different identities. This will lead to a better graph partitioning compared to the results by using the original noisy affinity graph.

\begin{figure}[t!]
\centering
\includegraphics[width=\textwidth]{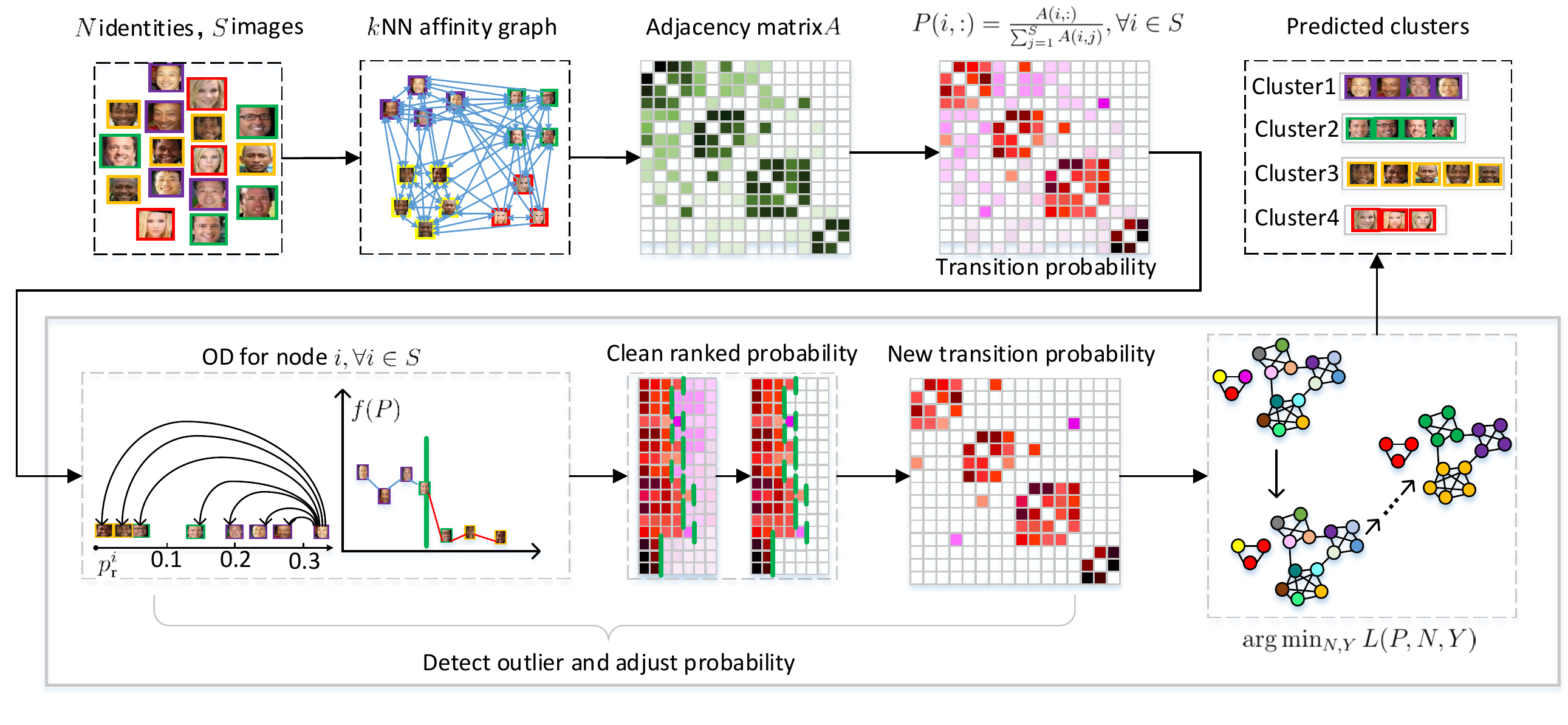}
\caption{The framework of FaceMap, where $p_{\text{r}}^i$ is the ranked non-zero transition probabilities of node $i$ and $f(P)$ is a  function transformation of $P$. We design a strategy of outlier detection adaptively adjusting the probabilities. By optimizing the map equation in Eq.~(\ref{eqn:face-loss}), the method effectively clusters facial images.}
\label{fig:framework}
\end{figure}

\subsection{FaceMap with Outlier Detection}

To adaptively detect the switch point in the ranked transition probabilities for each image, the key point is to discriminate the distribution change between the head and tail regions. Based on Observations 1-3 and Assumptions 1-2, we find that the switch point lies in the mixed region, and represents the outlier point of the stable distribution of the tail region.

The framework of FaceMap is demonstrated in Fig.~\ref{fig:framework}, where an outlier detection module is proposed based on $z$-score statistic of the average on $\Delta p_\text{r}^i$ over a window. With the detected outlier, new transition probabilities are updated and thus fed into Eq.~(\ref{eqn:face-loss}). For better illustrations, we show the algorithmic procedures in Algorithm~\ref{algo:adapt-infomap}. In FaceMap, we first get the ranked probabilities by descending order for each image. Then, we calculate the first-order sequence based on the ranked probabilities of each image. The module of outlier detection captures the switch point by maximizing $z$-score sequence. Finally, FaceMap adjusts transition probabilities and minimizes the entropy in Eq.~(\ref{eqn:face-loss}) with Infomap~\cite{rosvall2009map}.

Based on Assumptions 1-2, we construct a sliding window $\omega$ to calculate the average of $\Delta p_\text{r}^i$, \emph{i.e.}, $\hat{\mu}_j = (\sum^{j+\omega}_j \Delta p_\text{r}^i(j))/\omega$ with $j=k-\omega,\cdots,1$. Note that we start the sliding window from the end of the sequence of $\Delta p_\text{r}^i$. We introduce the sliding window because the switch point lies in the mixed region. In order to calculate the $z$-score, we estimate the mean and standard of the stable distribution of the tail region by $\bar{\mu}_j = (\sum^{k}_j \Delta p_\text{r}^i(j))/(k-j-1)$ and $\bar{\sigma}_j = \sqrt{\sum^{k}_j (\Delta p_\text{r}^i(j)-\bar{\mu}_j)^2/(k-j-1)}$ with $j=k-\omega,\cdots,1$. Note that in Algorithm~\ref{algo:adapt-infomap}, we set $k=K$. 

Without loss of generality, we mark the middle point of a sliding window as the candidate of outlier. We calculate the $z$-score of the sliding window as $z=|\hat{\mu}_j-\bar{\mu}_j|/\bar{\sigma}_j$, and maximize $z$-score over the sequence. We give the theoretical insight of maximizing the $z$-score. Based on Chebyshev-Bienayme's Inequality~\cite{seneta2013tricentenary}, we are ready to have the tail probability of the stable distribution as $\mathbf{P}\left[\frac{|\hat{\mu}_j-\bar{\mu}_j|}{\bar{\sigma}_j} > v\right] \leq \frac{1}{v^2}$, for all $v >0$. It means that the larger value of $z$-score of a point, the less likely it comes from the same distribution and thus the more possibility to be a switch point.

\begin{algorithm}[tb]
    \caption{FaceMap in pseudocode}
    \begin{algorithmic}[1]
    \STATE \textbf{input:} a set of face images with features $\{x_i\}^S_{i=1}$, $K$, window size $\omega$
    \STATE construct an affinity graph $A$ based on $k$NN with $k=K$
    \FOR{$i=1$ to $S$}
        \STATE $P(i, :) = A(i, :)/\sum^S_{j=1} A(i,j)$ \COMMENT{calculate the original transition probability}
        \STATE $p^i_\text{r} = \text{rank-d}(P(i,:))[:K]$ \COMMENT{rank $K$ probabilities by descending order}
        \FOR{$j=1$ to $K-1$}
        \STATE $\Delta p^i_\text{r}(j) = p^i_\text{r}(j) - p^i_\text{r}(j+1) $ \COMMENT{first-order values of the ranked probabilities}
        \ENDFOR
        \STATE initialize $z = \text{zeros}(K)$
        \FOR{$j=K-\omega-1$ to 1}
        \STATE $\hat{\mu}_j = \sum^{j+\omega}_j \Delta p_\text{r}^i(j)/\omega$ \COMMENT{mean of $\Delta p^i_\text{r}$ in the sliding window}
        \STATE $\bar{\mu}_j = \sum^{K-1}_j \Delta p_\text{r}^i(j)/(K-j-1)$ \COMMENT{mean of $\Delta p^i_\text{r}$ in the tail region}
        \STATE $\bar{\sigma}_j = \sqrt{\frac{\sum^{K-1}_j (\Delta p_\text{r}^i(j)-\bar{\mu}_j)^2}{K-j-1}}$ \COMMENT{standard deviation of $\Delta p^i_\text{r}$ in the tail region}
        \STATE $q = j + \lceil \omega/2 \rceil$ \COMMENT{mark the middle point as a candidate of outlier} 
        \STATE $z(q) = \frac{|\hat{\mu}_j - \bar{\mu}_j|}{\bar{\sigma}_j}$
        \ENDFOR
        \STATE $q^* = \arg \max_q z(q)$ 
        \COMMENT{maximize $z$-score to detect the outlier}
        \FOR{$j=1$ to $S$}
        \STATE $P(i, j) = P(i, j)>=p^i_\text{r}(q^*)?P(i, j):0$ \COMMENT{adjust transition probabilities}
        \ENDFOR
        
    \ENDFOR
    \STATE $N_\text{P}, Y_\text{P} = \arg\min_{N,Y} L(P, N,Y)$
    \STATE \textbf{output}: $Y_\text{P}$
\end{algorithmic}\label{algo:adapt-infomap}
\end{algorithm}

\section{Experiments}
In this section, we conduct experiments on three large-scale face datasets. The comparisons based on traditional metrics are shown via two categories, unsupervised and supervised settings. We further show the performance on new metrics among all methods. We also conduct ablation studies to demonstrate the consistent performance with respect to hyper-parameters of $k$ and $\omega$ in FaceMap, and the superiority of outlier detection in FaceMap for face clustering.

\subsection{Datasets and Metrics}
The three publicly available large-scale datasets of face images are MS-Celeb-1M~\cite{guo2016ms}, VGGFace2~\cite{cao2018vggface2} and CASIA~\cite{yi2014casiawebface}. Following~\cite{deng2019arcface}, we use the refined version of MS-Celeb-1M, denoted as MS1M.  The statistics of three datasets are shown in Table~\ref{table:stats}. From the table, we find that there is a large discrepancy of distribution of image numbers per identity among three datasets. It is challenging to solve face clustering problem across different datasets with good performance. 
\begin{table}[!t]
\centering
\caption{Statistics of datasets. AVG stands for average image numbers per identity. STD stands for standard deviation of image numbers per identity.}
\begin{tabular}{|c|c|c|c|c|}
\hline
     Datasets & \# of images & \# of identities & AVG & STD \\
     \hline
     MS1M~\cite{guo2016ms} & 5.8M & 85.0K & 68.5 & 40.6 \\
     \hline
     VGGFace2~\cite{cao2018vggface2} & 3.3M & 9.1K & 362.6 & 101.3\\
     \hline
     CASIA~\cite{yi2014casiawebface} & 0.5M &  10.5K & 46.4 & 59.3\\
     \hline
\end{tabular}
\label{table:stats}
\end{table}

For evaluation metrics, we first adopt the traditional $F_\text{P}$ and $F_\text{B}$ for all methods. Then, for evaluating identity-level quality of clustering, we calculate the values based on new metrics proposed in this paper, \emph{i.e.}, $R_{\#\text{I}}$, $R_{\#\text{S}}$ and $F_{\text{I}}$, which have been discussed in details in Section~\ref{sec:metrics}.

\subsection{Implementation Details}
Following the setting in~\cite{shen2021structure,yang2020learning,yang2019learning}, MS1M is divided into 10 parts with nearly equal number of identities. A face clustering test benchmark in multiple scales is built by combining different numbers of subsets. We use number of subsets and face numbers at multiple scales to denote these test sets, \emph{e.g.}, Part3(2.89M) denotes the union of 3 subsets of MS1M with 2.89M face instances. Similarly, We randomly select 10 percent of identities of each dataset as training set and the others as test set for VGGFace2 and CASIA. In the experiments, we take the training set to train the GCN-based models in supervised methods and evaluate on the test set for all the methods to make a fair comparison. We use extracted feature provided in~\cite{shen2021structure} as the face representation of MS1M. We use the pre-trained Arcface model, which is trained on Glint-360K~\cite{deng2019arcface,an2021partial}, to extract face representation features of VGGFace2 and CASIA. To build the affinity graph, we use Faiss~\cite{johnson2019billion} to search nearest neighbors for all methods. In experiments, we set $k$ as 256 for FaceMap and 80 for GCN-based methods. We set $\omega=20$.

\subsection{Results}
\textbf{1) Performance comparison with unsupervised methods.}
We report the clustering results in Table~\ref{tab:unsupervised_comparison}. FaceMap outperforms all the unsupervised methods by a large margin on all the test datasets. We note that the performance of most of the unsupervised methods is sensitive to the hyper-parameters, \emph{e.g.}, the pre-set cluster number in K-means. In addition, some methods are not scalable for large-scale datasets, \emph{e.g.}, HAC failed to cluster VGGFace2 in 360 hours. For ARO, we report superior results compared to the performance in~\cite{shen2021structure} by carefully tuning the thresholds.  The performance of DBSCAN is not consistent across the datasets as it assumes a balanced density in each cluster. Note that FaceMap has superior performance across different test datasets.



\begin{table}[!t]
    \centering
    \caption{Performance comparison with unsupervised methods. Our FaceMap yields gains of 10.38\% on $F_\text{P}$ and 5.75\% on $F_\text{B}$ on average, compared to the best results of unsupervised methods. We adopt $k=256$ and $\omega=20$ for FaceMap.}
    \resizebox{\columnwidth}{!}{\begin{tabular}{|c|c c|c c|c c|c c|c c|c c|c c|}
    \bottomrule
            \multicolumn{1}{|c|}{\textbf{Datasets}} & 
            \multicolumn{2}{c|}{\textbf{Part1(0.58M)}} & \multicolumn{2}{c|}{\textbf{Part3(1.74M)}} & \multicolumn{2}{c|}{\textbf{Part5(2.89M)}} & \multicolumn{2}{c|}{\textbf{Part7(4.05M)}} & \multicolumn{2}{c|}{\textbf{Part9(5.21M)}} &
            \multicolumn{2}{c|}{\textbf{CASIA}} & \multicolumn{2}{c|}{\textbf{VGGFace2}} \\
    \hline
    Methods &$F_\text{P}$ & $F_\text{B}$ &  $F_\text{P}$ & $F_\text{B}$ &  $F_\text{P}$ & $F_\text{B}$ &  $F_\text{P}$ & $F_\text{B}$ &  $F_\text{P}$ & $F_\text{B}$ &  $F_\text{P}$ & $F_\text{B}$ &  $F_\text{P}$ & $F_\text{B}$ \\
    \hline
    K-means~\cite{likas2003global} & 79.21 & 81.23 & 73.04 & 75.20 & 69.83 & 72.34 & 67.90 & 70.57 & 66.47 & 69.42 & 36.72 & 78.53 & 20.47 & 81.46 \\
    HAC~\cite{sibson1973slink} & 70.63 & 70.46 & 54.40 & 69.53 & 11.08 & 68.62 & 1.40 & 67.69 & 0.37 & 66.96 & 61.87 & 53.65 & NA & NA \\
    ARO~\cite{otto2017clustering} & 85.58 & 87.75 & 82.26 & 84.39 & 81.08 & 82.78 & 79.76 & 81.60 & 79.76 & 80.67 & 75.88 & 87.08 & 78.71 & 84.35 \\
    DBSCAN~\cite{ester1996density} & 67.93 & 67.17 & 63.41 & 66.53 & 52.50 & 66.26 & 45.24 & 44.87 & 44.94 & 44.74 & 57.25 & 49.43 & 66.88 & 65.49 \\
    \hline
     \multirow{2}{*}{\textbf{FaceMap}} & 94.24 & 92.55 & 91.31 & 89.67 & 89.32 & 88.20 & 87.74 & 87.11 & 86.37 & 86.29 & 92.55 & 91.24 & 94.15 & 93.78 \\
     & \textbf{+8.66} & \textbf{+4.80} & \textbf{+9.05} & \textbf{+5.28} & \textbf{+8.24} & \textbf{+5.42} & \textbf{+7.98} & \textbf{+5.51} & \textbf{+6.61} & \textbf{+5.62} & \textbf{+16.67} & \textbf{+4.16} & \textbf{+15.44} & \textbf{+9.43} \\
    \hline
    \end{tabular}}
    \label{tab:unsupervised_comparison}
\end{table}

\textbf{2) Performance comparison with supervised methods.} 
FaceMap achieves a better result on all the test datasets, as shown in Table~\ref{tab:gcn_comparison}. We directly report the results of MS1M in the related reference. The results of CASIA and VGGFace2 are produced using the source codes provided by the authors.
We observe that the performance of supervised methods on CASIA and VGGFace2 is not as good as the ones in MS1M. This gap is caused by the difference of labeled data and the generalization of the hyper-parameters.

\begin{table}[!t]
    \caption{Performance comparison with supervised methods. Our FaceMap yields gains of 4.80\% on $F_\text{P}$ and 5.04\% on $F_\text{B}$ on average, compared to the best results of supervised methods. We adopt $k=256$ and $\omega=20$ for FaceMap.}
    \centering
    \resizebox{\columnwidth}{!}{\begin{tabular}{|c|c c|c c|c c|c c|c c|c c|c c|}
    \bottomrule
            \multicolumn{1}{|c|}{\textbf{Datasets}} & 
            \multicolumn{2}{c|}{\textbf{Part1(0.58M)}} & \multicolumn{2}{c|}{\textbf{Part3(1.74M)}} & \multicolumn{2}{c|}{\textbf{Part5(2.89M)}} & \multicolumn{2}{c|}{\textbf{Part7(4.05M)}} & \multicolumn{2}{c|}{\textbf{Part9(5.21M)}} &
            \multicolumn{2}{c|}{\textbf{CASIA}} & \multicolumn{2}{c|}{\textbf{VGGFace2}} \\
    \hline
    Method &$F_\text{P}$ & $F_\text{B}$ &  $F_\text{P}$ & $F_\text{B}$ &  $F_\text{P}$ & $F_\text{B}$ &  $F_\text{P}$ & $F_\text{B}$ &  $F_\text{P}$ & $F_\text{B}$ &  $F_\text{P}$ & $F_\text{B}$ &  $F_\text{P}$ & $F_\text{B}$ \\
    \hline
    L-GCN~\cite{wang2019linkage}~CVPR~2019 & 78.68 & 84.37 & 75.83 & 81.61 & 74.29 & 80.11 & 73.70 & 79.33 & 72.99 & 78.60 & 68.52 & 79.22 & 60.74 & 66.42 \\
    GCN-D~\cite{yang2019learning}~CVPR~2019  & 85.66 & 85.52 & 83.76 & 83.99 & 81.62 & 82.00 & 80.33 & 80.72 & 79.21 & 79.71 & 78.71 & 83.14 & 65.80 & 69.27\\
    GCN-V+E~\cite{yang2020learning}~CVPR~2020 & 87.93 & 86.09 & 84.04 & 82.84 & 82.10 & 81.24 & 80.45 & 80.09 & 79.30 & 79.25 & 83.40 & 81.35 & 50.35 & 52.47\\
    Clusformer~\cite{nguyen2021clusformer}~CVPR~2021 & 88.20 & 87.17 & 84.60 & 84.05 & 82.79 & 82.30 & 81.03 & 80.51 & 79.91 &  79.95 & NA & NA & NA & NA \\
    CPC~\cite{liu2021learn}~ICCV~2021 & 90.67 & 89.54 & 86.91 & 86.25 & 85.06 & 84.55 & 83.51 & 83.49 & 82.41 & 82.40 & NA & NA & NA & NA \\
    STAR-FC~\cite{shen2021structure}~CVPR~2021 & 91.97 & 90.21 & 88.28 & 86.26 & 86.17 & 84.13 & 84.70 & 82.63 & 83.46 & 81.47 & 75.34 & 71.81 & 84.12 & 83.49 \\
    \hline
     \multirow{2}{*}{\textbf{FaceMap}} & 94.24 & 92.55 & 91.31 & 89.67 & 89.32 & 88.20 & 87.74 & 87.11 & 86.37 & 86.29 & 92.55 & 91.24 & 94.15 & 93.78 \\
     & \textbf{+2.27} & \textbf{+2.34} & \textbf{+3.03} & \textbf{+3.41} & \textbf{+3.15} & \textbf{+3.65} & \textbf{+3.04} & \textbf{+3.62} & \textbf{+2.91} & \textbf{+3.89} & \textbf{+9.15} & \textbf{+8.10} & \textbf{+10.03} & \textbf{+10.29} \\
    \hline
    \end{tabular}}
    \label{tab:gcn_comparison}
\end{table}

\textbf{3) Performance comparison of identity-level quality.}
We compare our methods from the perspective of identity-level quality in Table~\ref{tab:idf1}. Note that STAR-FC (which achieves the state-of-the-arts in terms of $F_\text{B}$ and $F_\text{P}$) is even worse than the unsupervised method ARO in light of the new metrics. This implies that the traditional metrics have missed the identity-level evaluation. We can also observe that, compared to all existing methods, $R_{\#\text{I}}$ is closer to 100\% and $R_{\#\text{S}}$ is much smaller in FaceMap. This result demonstrates that FaceMap has a great advantage to estimate the true number of identity. In addition, our method yields a large boost on $F_{\text{I}}$ with $\theta$=0.5 and 0.9, which demonstrates that a large number of identities with high quality clusters are generated by FaceMap.
 
 \begin{table}[!tp]
    \centering
    \caption{Identity-level comparison via new metrics. FaceMap results high quality clusters at identity level compared to other methods. All the results are reported in $\%$. $\downarrow$ indicates the smaller results are better.}
    \resizebox{.8\columnwidth}{!}{\begin{tabular}{|c|c c c c|c c c c|c c c c|}
    \bottomrule
         \multicolumn{1}{|c|}{\textbf{Datasets}} & \multicolumn{4}{c|}{\textbf{CASIA}} & \multicolumn{4}{c|}{\textbf{VGGFace2}} &\multicolumn{4}{c|}{\textbf{Part1(0.58M)}} \\
    \hline
        Methods  & $F_{\text{I}}$(0.5) & $F_{\text{I}}$(0.9) & $R_{\#\text{S}}$($\downarrow$) & $R_{\#\text{I}}$ & $F_{\text{I}}$(0.5) & $F_{\text{I}}$(0.9) & $R_{\#\text{S}}$($\downarrow$) & $R_{\#\text{I}}$  & $F_{\text{I}}$(0.5) & $F_{\text{I}}$(0.9) & $R_{\#\text{S}}$($\downarrow$) & $R_{\#\text{I}}$ \\
    \hline
    HAC~\cite{sibson1973slink} & 4.45 & 0.28 & 1764.75 & 1990.40 & NA & NA & NA & NA & 8.83 & 2.13 & 896.29 & 1432.17 \\ 
    ARO~\cite{otto2017clustering} & 32.36 & 22.11 & 348.01 & 505.03 & 4.88 & 1.29 & 321.66 & 3953.92 & 26.18 & 14.51 & 372.17 & 596.36 \\
    DBSCAN~\cite{ester1996density} & 3.60 & 0.19 & 1917.78 & 2151.39 & 1.68 & 0.02 & 9013.06 & 10019.62 & 7.64 & 1.63 & 981.63 & 1571.83 \\
    \hline
    L-GCN~\cite{wang2019linkage} &17.19 &6.56 &694.09 &807.68 &1.37 &0.42 &11424.99 &11667.92 &23.70 &14.32 &393.83 &519.95 \\
    GCN-D~\cite{yang2019learning} &28.65 &14.01 &308.87 &508.90 &2.58 &0.51 &6253.86 &7002.88 &34.12 &19.22 &216.61 &437.94 \\
    GCN-V+E~\cite{yang2020learning} &21.96 &7.97 &444.68 &624.06 &6.11 &0.39 &615.82 &1508.31 &19.78 &11.40 &409.08 &665.27 \\
    STAR-FC~\cite{shen2021structure} & 7.89 & 1.93 & 1281.97 & 1387.47 & 2.20 & 1.55 & 7313.88 & 7413.10 & 23.72 & 17.62 & 466.91 & 574.72 \\
    \hline
    \textbf{FaceMap} & \textbf{92.45} & \textbf{66.30} & \textbf{3.15} & \textbf{110.88} & \textbf{63.99} & \textbf{51.88} & \textbf{26.55} & \textbf{209.85} & \textbf{89.05} & \textbf{67.53} & \textbf{1.94} &  \textbf{110.77} \\
    \hline
    \end{tabular}}
    \label{tab:idf1}
\end{table}
 

\subsection{Ablation Studies}

\textbf{1) Study on the sensitivity of $K$ and $\omega$.}  FaceMap has two hyper-parameters, \emph{i.e.}, $K$ and $\omega$. We note that $K$ affects the construction of the affinity graph, and $\omega$ can influence the detection of the outlier position. Table~\ref{tab:ablatoin_windowsize} shows the performance of FaceMap when we vary $K$ and $\omega$ on the MS1M Part1(0.58M).  We observe that FaceMap exhibits stable performance on all the metrics except $R_{\#\text{S}}$, and larger $K$ and smaller $\omega$ result in smaller $R_{\#\text{S}}$. This result shows that FaceMap is robust.

\begin{table}[!t]
    \centering
    \caption{Performance of FaceMap on MS1M Part1(0.58M) with different $K$ and $\omega$. All the results are reported in $\%$. $\downarrow$ indicates the smaller results are better.}
     \resizebox{1.0\linewidth}{!}{
    \begin{tabular}{|c|c|c|c|c|c|c|c|c|c|c|c|c|c|c|c|c|c|}
    \hline
    $K$ & \multicolumn{5}{c|}{128} & \multicolumn{5}{c|}{256} & \multicolumn{5}{c|}{512} & \multirow{2}{*}{\textbf{Mean}} & \multirow{2}{*}{\textbf{STD}} \\
    \cline{1-16}
    $\omega$ & 10 & 15 & 20 & 25 &30 & 10 & 15 & 20 & 25 &30 & 10 & 15 & 20 & 25 &30 & & \\
    \hline
    $F_\text{P}$ &94.02 &94.21 &94.36 &94.39 &94.32 &94.21 &94.23 &94.24 &94.16 &94.01 &94.02 &93.99 &93.85 &93.84 &93.57 &94.10 &0.22  \\
    \hline
    $F_\text{B}$ &92.31 &92.50 &92.61 &92.63 &92.58 &92.54 &92.56 &92.55 &92.51 &92.37 &92.42 &92.41 &92.31 &92.28 &92.06 &92.44 &0.15 \\
    \hline
    $F_{\text{I}}$(0.9) &64.33 &65.41 &66.49 &67.03 &67.40 &66.41 &67.23 &67.53 &67.58 &67.97 &66.50 &67.25 &67.48 &67.60 &67.74 &66.93 &0.95 \\
    \hline
    $R_{\#\text{S}}$($\downarrow$) &1.99 &2.25 &2.89 &3.10 &3.01 &1.35 &1.54 &1.94 &2.16 &1.98 &1.00 &1.19 &1.40 &1.66 &1.33 &1.92 &0.64 \\
    \hline
    $R_{\#\text{I}}$ &120.32 &118.06 &115.22 &113.90 &111.79 &115.19 &112.99 &110.77 &109.46 &106.95 &113.18 &110.71 &107.85 &106.89 &104.41 &111.85 &4.23 \\
    \hline
    \end{tabular}}
    \label{tab:ablatoin_windowsize}
\end{table}


\begin{figure}[!t]
    \centering
    \begin{subfigure}{.5\textwidth}
    \includegraphics[width=.9\linewidth]{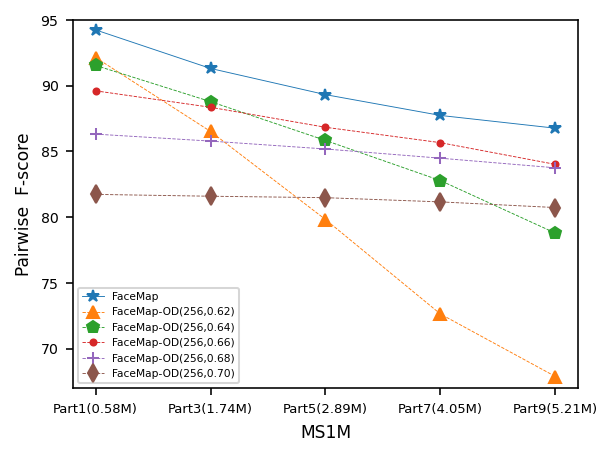}
    \end{subfigure}%
    \begin{subfigure}{.5\textwidth}
    \includegraphics[width=.9\linewidth]{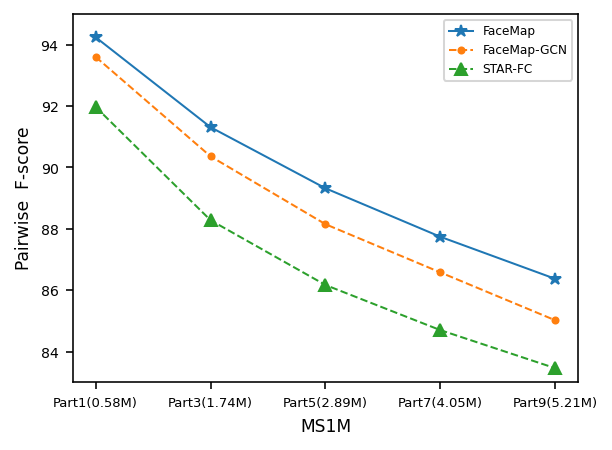}
    \end{subfigure}\\
    \begin{subfigure}{.5\textwidth}
    \includegraphics[width=.9\linewidth]{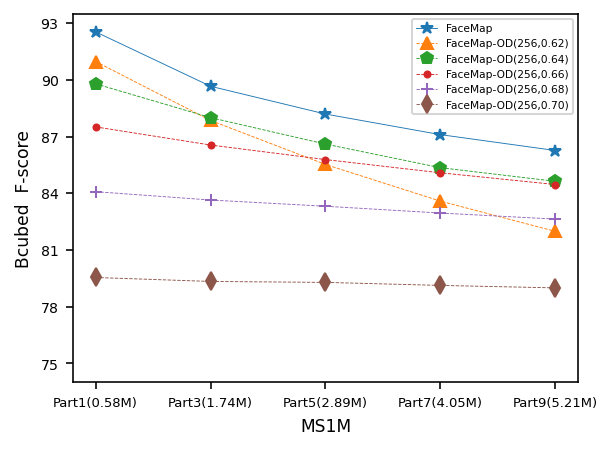}
    \end{subfigure}%
    \begin{subfigure}{.5\textwidth}
    \includegraphics[width=.9\linewidth]{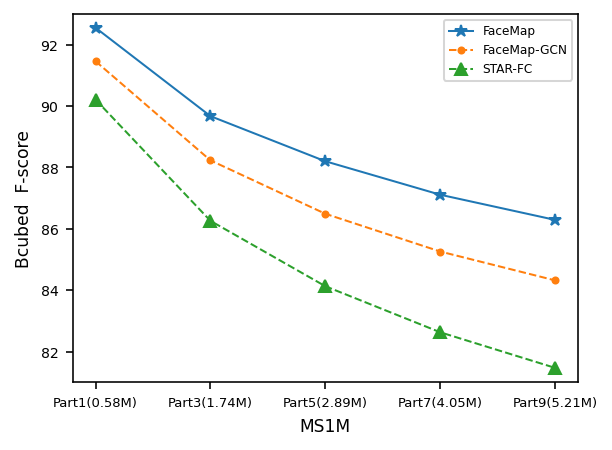}
    \end{subfigure}%
    \caption{Ablation study on the effectiveness of outlier detection based on $F_\text{P}$ and $F_\text{B}$.}
    \label{fig:ablation_od}
\end{figure}

\textbf{2) Study on the effectiveness of outlier detection.} We denote FaceMap without the OD module by FaceMap-OD. We introduce hand-crafted rules to reduce the noisy edges in the affinity graph by fixing a similarity threshold $\delta$ based on $k$NN,  represented by FaceMap-OD($k$,$\delta$). In addition, we delete the OD module and adopt a learned affinity graph from GCN-based method (STAR-FC) as input for the map equation in Eq.~(\ref{eqn:face-loss}), which is represented by FaceMap-GCN. We show the ablation study on the effectiveness of the OD module based on $F_\text{P}$ and $F_\text{B}$ in Fig.~\ref{fig:ablation_od}. We can observe that FaceMap consistently outperforms all other methods with different thresholds, which demonstrates the effectiveness and robustness of the outlier detection module across multiple scales of datasets. Besides, we observe that FaceMap-GCN performs better than STAR-FC, and FaceMap exhibits a consistent better performance compared to FaceMap-GCN. This result further validates the superiority of the proposed outlier detection module.

\section{Conclusions}

In this paper, we propose an unsupervised method, named as FaceMap, for large-scale face clustering. We formulate face clustering as community detection with the map equation. By minimizing the map equation on the entropy of the structure in affinity graph from facial images, we obtain the predicted clusters of facial images. In order to alleviate the noisy transition probability, we develop a strategy of outlier detection to adaptively adjust transition probabilities. We also illustrate the limitations of the traditional metrics for face clustering and design three metrics for comprehensive evaluations. The proposed FaceMap achieves new state-of-the-arts in light of traditional metrics on three large-scale datasets, where our method significantly outperforms the prior methods. We demonstrate the superiority of FaceMap in identity-level quality via new metrics. From a viewpoint of practical applications with super large-scale datasets, we may take the advantage of the distributed characteristic in calculation of outlier detection module for FaceMap, and reduce the computational complexity with a high-performance distributed system.  Further works on this direction lie in a broader applications of the map equation to more computer vision tasks.



\clearpage
%
%
\bibliographystyle{plain}
\bibliography{egbib}
\end{document}